\documentclass{article}

\usepackage{microtype}
\usepackage{graphicx}
\usepackage{booktabs}
\usepackage{hyperref}

\usepackage[accepted]{icml2026}

\ifdefined\LOCALBUILD\fi

\makeatletter
\renewcommand{\ICML@appearing}{\textit{ICML 2026 Workshop on AI for Law
(AI4Law)}, Seoul, South Korea, 2026. Copyright 2026 by the author(s).}
\makeatother

\usepackage{amsmath}
\usepackage{amssymb}
\usepackage{eurosym}        
\usepackage{xcolor}
\usepackage{tikz}           
\usetikzlibrary{arrows.meta, positioning}

\icmltitlerunning{Temporal Misgrounding in Legal RAG}

\begin{document}

\twocolumn[
\icmltitle{Temporal Misgrounding in Legal RAG:\\
A Versioned-Corpus Benchmark for French Tax Law}

\begin{icmlauthorlist}
\icmlauthor{Rose Cymbler}{talia}
\icmlauthor{Daniel Guez}{talia}
\icmlauthor{Laurent Fabre}{databricks}
\end{icmlauthorlist}
\icmlaffiliation{talia}{Talia, Paris, France}
\icmlaffiliation{databricks}{Databricks}
\icmlcorrespondingauthor{Rose Cymbler}{rose.cymbler@talia-ai.com}
\icmlkeywords{legal NLP, retrieval-augmented generation, temporal grounding, benchmark}
\vskip 0.3in
]

\printAffiliationsAndNotice{\textbf{Author contributions:} R.C. led the research,
designed the benchmark, built the retrieval architecture, ran the
controlled experiment, and led the writing. D.G. built the data infrastructure,
extraction, and versioned corpus. L.F. contributed to the evaluation methodology, experimental design, analysis and interpretation of results, and the positioning of the work. }

\begin{abstract}
We identify and quantify \textbf{temporal misgrounding}: the systematic retrieval and citation of the \emph{currently in-force} version of a legal article when the applicable version is an earlier or future one. Standard legal RAG treats the corpus as static; we argue legal question answering is a \emph{temporally-indexed} retrieval problem. We introduce \textbf{FiscalQA Pro}, pairing a versioned corpus of 32{,}436 article-versions of the French tax code (93 years, 1938--2031) with an all-model-hard temporal-reasoning track: 209 scored, expert-reviewed questions across 33 CGI articles (221 released; twelve flagged out of the answerable scope). At selection time, no evaluated model recovered its date-applicable answer closed-book in any of four sampling draws, and the currently in-force text lacks the gold value for all but one of the scored questions. Answers are scored \emph{deterministically} via atomic ground-truth ``nuggets'' (regex and numeric-with-tolerance), never LLM-as-judge: an LLM judge would inherit the temporal bias it is meant to score.
Across eleven models (five frontier closed-API systems plus Gemini 2.5 Pro as a substitute entry, and five open-weight), parametric knowledge yields 3.0\% mean strict accuracy and RAG over a static current-version corpus 2.7\%. Static RAG retrieves the date-applicable version 0\% of the time, confidently citing a real but inapplicable version. Our end-to-end retriever over a multi-version index, with no oracle, reaches \textbf{98.3\% mean strict}; an oracle-article ablation reaches 99.1\%, locating the residual gap in first-stage recall, not version selection. We additionally release a version-aware jurisprudence dataset of 69{,}208 citation links, together with the corpus, benchmark, model responses, and pipeline code.\footnote{\url{https://github.com/rosecymbler/fiscal-fr-bench}}
\end{abstract}

\section{Introduction}
\label{sec:intro}

\paragraph{Motivation.}
Recent progress in large language models (LLMs) and retrieval-augmented generation (RAG) has spurred a wave of legal-NLP benchmarks aimed at measuring legal reasoning, citation extraction, and document understanding \citep{guha2024legalbench, niklaus2023lextreme, douka2021juribert}. Yet most of these benchmarks share a critical implicit assumption: that the legal corpus is \emph{static}. This assumption is most dramatically violated in tax law, where statutes are amended yearly through finance laws (\textit{lois de finances}) and where the correct answer to a legal question often depends critically on the date of application. A taxpayer asking ``what was the standard corporate income tax rate in 2018?'' should receive \emph{$33\tfrac{1}{3}\%$}, not the current \emph{25\%}. Yet a frontier LLM trained through 2025, with or without retrieval over a current-version corpus, will reliably return the latter.

\paragraph{Phenomenon: temporal misgrounding.}
We give a name to this failure mode: \textbf{temporal misgrounding}. It is the systematic substitution of the currently in-force version of a legal article for the version that should be applied given the temporal context of the question. Temporal misgrounding has three concurrent root causes:
\begin{enumerate}
  \item \emph{Parametric recency bias.} LLMs are trained on snapshots that overrepresent the most recent legal state, biasing their priors toward current law.
  \item \emph{Absence of temporal conditioning in retrieval.} Standard dense retrievers index a corpus by semantic similarity without conditioning on the date implicit in the query.
  \item \emph{Article-number aliasing.} Article numbers persist across versions (e.g., article 219 of the CGI exists in every version of the code), so naive retrieval cannot distinguish which version is intended.
\end{enumerate}

\paragraph{Research question.}
We ask: \emph{On legal questions whose date-applicable answer differs from the currently in-force law, to what extent does temporal misgrounding bottleneck the accuracy of state-of-the-art LLMs and RAG systems?} We isolate this regime deliberately (\S\ref{sec:benchmark}): it is where temporal misgrounding is diagnostic, so our reported gaps are conditional on it rather than an average over all legal QA. We hypothesize that (H) temporal validity is a first-order axis of legal grounding, and that explicitly conditioning retrieval on the temporal context of the query closes most of the gap between LLM-only and version-conditioned performance, and that an end-to-end retriever, not only an oracle, realizes this.

\paragraph{Contributions.}
We support this hypothesis with the following contributions:

\begin{itemize}
  \item \textbf{A characterization of temporal misgrounding} as a distinct failure mode of legal RAG, with a taxonomy of four sub-modes (\S\ref{sec:problem}).
  \item \textbf{A versioned corpus} of 32{,}436 CGI/LPF article-versions spanning 93 years (1938--2031), including future-effective versions, plus an auxiliary dataset of 69{,}208 version-aware jurisprudence links (98--99\% precision; \S\ref{sec:corpus}, App.~\ref{app:linking}).
  \item \textbf{A benchmark} whose R3 temporal track is 209 scored, expert-reviewed, all-model-hard questions across 33 CGI articles (221 released) in a four-regime framework, scored deterministically via nuggets rather than LLM-as-judge (\S\ref{sec:benchmark}); R2/R4 released as additional tracks.
  \item \textbf{A controlled three-condition experiment} (\S\ref{sec:experiments}) quantifying the LLM-only / static-RAG / version-aware-RAG accuracy gap on temporal questions.
\end{itemize}

\paragraph{Positioning.}
FiscalQA Pro extends the methodology of OfficeQA Pro \citep{opsahlong2026officeqa} (deterministic grounded reasoning over U.S.\ Treasury Bulletins) to a setting where the underlying documents are revised \emph{in place} rather than appended, and where the date of application is itself a first-class retrieval signal. Unlike SAT-Graph RAG \citep{demartim2025satgraph}, which resolves point-in-time legal queries through an ontology-driven knowledge graph, our method needs no ontology or graph construction, only explicit version indexing and date-conditioned retrieval over the raw versioned corpus. We focus on tax law because it maximizes the phenomenon under study: yearly amendments through finance laws, date-sensitive answers (rates, thresholds, deduction rules), and richly cross-referenced jurisprudence enabling version-aware linking of statutes to case law.

\section{Related Work}
\label{sec:related}

\paragraph{Legal benchmarks.}
LegalBench \citep{guha2024legalbench} provides 162 tasks covering U.S.\ legal reasoning across rule application, issue spotting, and rhetorical understanding, all in English and predominantly without temporal indexing. LEXTREME \citep{niklaus2023lextreme} extends multi-lingual coverage with 18 tasks spanning 24 languages across multiple legal systems. LEXam \citep{fan2026lexam} benchmarks legal reasoning on 340 bilingual (EN/DE) law exams across multiple jurisdictions, with rubric-based scoring; orthogonal to our axis, it evaluates reasoning quality on a fixed legal state rather than temporal grounding across versions.
JuriBERT \citep{douka2021juribert} is a French legal BERT, and ClaimRAG-LAW \citep{das2026claimrag} a fine-grained claim-level legal RAG benchmark. None of these benchmarks explicitly evaluate temporal drift in legal QA.

\paragraph{Enterprise grounded reasoning.}
OfficeQA Pro \citep{opsahlong2026officeqa} introduced an enterprise benchmark for end-to-end grounded reasoning over U.S.\ Treasury Bulletins, comprising 133 questions across an 89{,}000-page corpus spanning nearly a century. Their evaluation is deterministic: answers are checked against numerical values, citations, and dates rather than via LLM-as-judge. They show that frontier LLMs achieve below 5\% accuracy on parametric knowledge alone, 12\% with web access, and 34.1\% with direct corpus access (signaling substantial headroom even at the frontier).

\paragraph{Agentic retrieval and nugget-based scoring.}
KARL \citep{databricks2026karl} introduces KARLBench, a multi-capability evaluation suite spanning six search regimes, and proposes a \emph{nugget-based} metric: ground-truth answers are converted into atomic ``nuggets,'' and a model's response is scored by the fraction of nuggets it covers. We note that nugget scoring as instantiated in KARL remains an LLM-as-judge variant: an LLM judges whether each nugget is supported by the response, but one that anchors the judgment in expert-curated atomic facts rather than free-form rubrics. They also introduce OAPL, an off-policy reinforcement-learning procedure for training enterprise search agents. We adopt their nugget decomposition and multi-task framing, but remove the judge entirely by restricting nuggets to deterministically checkable types: article identifiers matched by regex and numeric values matched with tolerance (\S\ref{ssec:nuggets}).

\paragraph{Temporal grounding in NLP.}
Temporal question answering is surveyed by \citet{piryani2025hightime}. TempLAMA \citep{dhingra2022templama} and TimeQA \citep{chen2021timeqa} explore temporal reasoning in factual QA, finding that LLMs systematically default to the most recent fact seen in training. \citet{vu2023freshqa} extend this to dynamic open-domain QA with their FreshQA benchmark. LexTime \citep{barale2025lextime} studies temporal event-ordering in U.S.\ federal complaints; FiscalQA Pro complements this work by addressing version-conditioning of statutory retrieval (where the temporal axis is the law itself rather than the events under adjudication). Most directly, SAT-Graph RAG \citep{demartim2025satgraph} resolves point-in-time legal queries through an ontology-driven knowledge graph (contrasted with our graph-free approach in \S\ref{sec:intro}), but evaluates it as a qualitative case study on the Brazilian Constitution rather than a scored benchmark. Two works concurrent with our submission study the same failure: \citet{fan2026timetravel} diagnose a training-cutoff bias and the absence of temporal constraints in legal search agents (our causes (1)--(2)) over a 13-task RL benchmark, and \citet{prior2026oldfriend} release 312 expert-validated German statutory QA pairs with version-filtering RAG conditions. Their independent convergence underscores that temporal misgrounding is a real, general failure of legal RAG. We differ on three axes: our scoring is deterministic (regex / numeric-with-tolerance), whereas \citet{prior2026oldfriend} use an LLM-as-judge that inherits the very temporal bias it scores (\S\ref{ssec:nuggets}); our corpus is version-indexed at fine granularity (32{,}436 versions over 93 years, incl.\ future-effective); and our all-model-hard filter spans eleven models. We intend \emph{temporal misgrounding} as an umbrella subsuming their post-cutoff staleness and recency bias.

\paragraph{French legal NLP.}
Available resources include the open Légifrance API (DILA/PISTE) for legislation, Judilibre for Cour de cassation jurisprudence, and Arianeweb for Conseil d'État decisions. We build directly on these public sources, with full provenance preserved at the article-version level.

\section{Why Static RAG Fails on Legal QA}
\label{sec:problem}

Before describing the corpus and benchmark, we articulate why standard RAG architectures (which dominate current legal AI deployments) systematically fail on temporally-grounded legal questions. We argue that the root cause is structural, not anecdotal: it arises from the conjunction of three independent properties of (i) the legal corpus, (ii) typical retrieval architectures, and (iii) LLM parametric priors.

\subsection{Structural Properties of Legal Corpora That Defeat Static Retrieval}
\label{ssec:structural}

\paragraph{(P1) Same identifier, different content.}
Article 219 of the French CGI exists at every point in time, but its substantive content (e.g., the standard corporate income tax rate) differs across versions: \emph{$33\tfrac{1}{3}\%$} in 2018, \emph{31\%} in 2019, \emph{28\%} in 2020, \emph{26.5\%} in 2021, \emph{25\%} from 2022 onward. Article identifiers (\texttt{cid}) are stable, but \texttt{(article\_id, version\_id)} pairs vary. A retriever indexing only one snapshot of the corpus collapses this fine-grained structure into a single point in retrieval space.

\paragraph{(P2) Date-dependent correctness.}
Each version carries explicit \texttt{date\_debut}/\texttt{date\_fin} fields, so a question anchored in 2018 has a single correct version, distinct from 2020 or 2024. Unlike encyclopedic corpora, legal correctness is binary and conditional on date: a ``related'' or ``approximately correct'' version is simply wrong.

\paragraph{(P3) Cross-version semantic similarity.}
Successive versions of an article are textually near-identical, often differing by a single rate or threshold. Their dense embeddings are mutual nearest neighbors under any standard sentence encoder, making them nearly impossible to disambiguate by similarity alone.

\subsection{A Taxonomy of Temporal Misgrounding Failure Modes}
\label{ssec:taxonomy}

The interaction of (P1)--(P3) with current RAG architectures produces four failure modes: \textbf{current-law substitution} (the retriever returns the in-force version regardless of the query's date: the dominant mode, from a current-only index or recency-weighting); \textbf{future-law leakage} (a not-yet-in-force \texttt{VIGUEUR\_DIFF} version is returned for a present-tense question); \textbf{wrong-amendment resolution} (a query for an article ``in its version prior to law X'' yields the current or an arbitrary historical version; common in jurisprudence, where courts apply a deprecated version to facts arising under it); and \textbf{multi-version confusion} (a before/after-reform comparison collapses to the single highest-scoring version). We observe the first consistently across frontier LLMs; the other three require question types outside our single-anchor R3 set (\S\ref{sec:future}).

\subsection{Why This Is Not Solved by Larger or More Recent LLMs}
\label{ssec:not-solved}

One might expect newer LLMs, trained on larger legal corpora, to solve temporal misgrounding parametrically. Three observations argue otherwise. (i) Recency bias is structural: training data skews toward recently-published, widely-cited content, overrepresenting current law. (ii) The historical volume is enormous (30k+ versions across CGI and LPF alone), and memorizing it does not solve the disambiguation problem (P3). (iii) Temporally-grounded questions are easy to \emph{recognize} but hard to \emph{answer} without a version-indexed corpus, as our parametric knowledge filter shows empirically (\S\ref{sec:benchmark}; \citealp{opsahlong2026officeqa}). Temporal grounding thus requires structural changes to retrieval (version indexing, date-conditioning), not bigger models. We formalize the resulting hypotheses in \S\ref{sec:experiments}, each a retrieval condition we evaluate.

\section{Corpus Construction}
\label{sec:corpus}

We construct the FiscalQA Pro corpus in two stages: (i) extraction of temporally-versioned tax legislation, and (ii) version-aware linking of tax-related court decisions to the relevant article versions. All extraction is fully reproducible from public sources via the scripts released with this paper.

\subsection{Temporally-Versioned Tax Legislation}

\paragraph{Source.} The CGI (\textit{Code général des impôts}), its four annexes, and the LPF (\textit{Livre des procédures fiscales}) are extracted from the official PISTE Légifrance API (DILA): the \texttt{/consult/getArticleByCid} endpoint returns an article's complete version history (vs.\ only the current version from \texttt{/consult/getArticle}), each version carrying a distinct identifier (\texttt{LEGIARTI...}) under a stable \texttt{cid}.

\paragraph{Pipeline.} For each LEGITEXT code, we enumerate the constant identifiers via the table-of-contents endpoint, then call \texttt{getArticleByCid} once per \texttt{cid} and upsert every returned version with its \texttt{date\_debut}, \texttt{date\_fin}, \texttt{etat} (in force, modified, repealed, future-effective), and full text. The pipeline is idempotent and fully reproducible from public sources.

\paragraph{Output.} Our final corpus comprises 32{,}436 article-versions across six tax codes (per-code breakdown in App.~\ref{app:corpus}, Table~\ref{tab:corpus-stats}). The CGI corpus alone contains 21{,}040 versions over 3{,}696 distinct articles, an average of 5.69 historical versions per article (top: 94 versions for article 81, on tax-exempt income). The temporal span runs from 1938 (LPF) to 2031 (CGI articles whose entry into force has been deferred by recent finance laws), totaling 93 years.

\subsection{Version-Aware Jurisprudence Linking}
\label{ssec:linking}

As a secondary resource, we link tax-related court decisions to the \emph{specific article version} applicable at the decision date: a regex extractor (proximity veto + fiscal-context filter) resolves each citation against the article index and intersects it with \texttt{date\_debut}/\texttt{date\_fin} ranges. This yields \textbf{69{,}208 version-aware links across 32{,}034 decisions}, with \textbf{98--99\% link-level precision} (stratified audit) and 83--93\% decision-level recall on jurisdictional sources. Auxiliary to the controlled experiment, it serves as weak supervision for the R1/R4 tracks; full construction and audit are in Appendix~\ref{app:linking}.

\section{Benchmark Design}
\label{sec:benchmark}

FiscalQA Pro defines four question regimes (R1--R4) inspired by KARLBench's multi-task framing. This paper's benchmark and controlled experiment center on the \textbf{R3 temporal-reasoning track: 209 scored, expert-reviewed, all-model-hard questions across 33 CGI articles} (221 curated questions are released; twelve are flagged out of the answerable scope and excluded from scoring: four whose date-applicable value is an annually INSEE-indexed figure published administratively, conservatively excluded at curation, and eight excluded in a full-set review as curation errors or ill-posed under their date anchor), each paired with atomic ground-truth \textbf{nuggets} that enable deterministic scoring without LLM-as-judge. Regimes R2 (43 questions) and R4 (209 questions) are released as additional evaluation tracks.

\subsection{Question Regimes}

\paragraph{R1 -- Citation Extraction.}
Given a court decision and a legal question, identify the CGI/LPF article(s) grounding the court's reasoning; scored by exact match against the cited \texttt{cid} (e.g., a 2023 Conseil d'État decision $\rightarrow$ \texttt{LEGIARTI000006303451}, art.~209 B CGI).

\paragraph{R2 -- Deterministic Computation.}
Apply a tax rate or rule to a numerical case, scored by exact match (with tolerance): e.g., corporate income tax on a \euro{}1.2M FY2024 profit under the standard regime $\rightarrow$ \euro{}300{,}000 ($1.2\text{M}\times 25\%$).

\paragraph{R3 -- Temporal Reasoning (key contribution).}
Questions whose correct answer depends on the version of the law applicable at a specific date. Evaluation: two required nuggets identifying the article and the exact numerical value at that date, augmented by a retrieval-side provenance check (\S\ref{ssec:nuggets}). This is the regime targeted by our controlled experiment (Section~\ref{sec:experiments}).
\begin{quote}
\textit{``What was the standard rate of corporate income tax applicable to fiscal years opening on or after 1 January 2018, under the general regime (excluding SMEs)?''}\\
Expected nuggets: \texttt{art\_219\_CGI} (regex \texttt{\textbackslash b219\textbackslash b}) and \texttt{33.333\%} (numeric with tolerance); the applicable version is validated against retrieval provenance.
\end{quote}
We use article 219 here, and in the case study of App.~\ref{app:case-study}, for clarity; the \emph{scored} set deliberately targets less-memorized articles (Table~\ref{tab:distribution}, and the parametric filter below).

\paragraph{R4 -- Multi-Document Synthesis.}
Combine CGI, BOFiP (administrative doctrine), and case law to answer a complex question, scored by nuggets (3--7 per question spanning article, rate, doctrine reference, and a grounding decision).

\subsection{Nugget-Based Scoring}
\label{ssec:nuggets}

Following \citet{databricks2026karl}, we score a response $r$ by the fraction of ground-truth nuggets it covers, $\mathrm{score}(r,N)=\frac{1}{|N|}\sum_{i}\mathbf{1}[n_i\in r]$, where a nugget is an atomic, deterministically-checkable claim (article identifier, numerical value) and $\mathbf{1}[\cdot]$ is a regex match (identifiers), numeric-with-tolerance match (values, after arithmetic normalization: FR thousands separators, decimal comma), or exact string match, \emph{never} an LLM judgment. We additionally report a \emph{strict} score (both required nuggets hit: correct article and correct value) and a \emph{provenance} score for retrieval conditions: the date-applicable gold version is among the retrieved versions (the single article for B and $C_{or}$, the top-5 for $C_{prod}$).

\paragraph{Article-nugget leakage.} The article nugget is largely given away by the question text itself: its regex fires on the \emph{prompt} for 175 of the 209 scored questions (83.7\%), so citing the article is often not evidence of retrieval, and coverage is dominated by the value nugget. We therefore also report \emph{value-only coverage} (the value nugget alone), which tracks strict accuracy to within 1 point pooled (A 3.0\%, B 2.7\%, $C_{or}$ 99.2\%, $C_{prod}$ 98.5\%; per-model table in the repository's \texttt{stats\_clustered\_report.md}); the headline effect is carried entirely by the date-anchored value. On the retrieval side, the same fact means $C_{prod}$ receives the article number in 83.7\% of queries: the article-recall figures of \S\ref{ssec:killer-results} are conditioned on that information being present, as they would be in practice for a lawyer's query.

\paragraph{Why nuggets, not LLM-as-judge?}
Temporal grounding is precisely the axis on which frontier LLMs share a systematic bias (\S\ref{ssec:not-solved}), so an LLM judge inherits the failure mode it is asked to score: it can accept a fluent answer that quotes the wrong-date value because that value matches its own parametric prior. Deterministic nuggets sidestep this circularity, at the cost of narrower expressivity (a nugget is atomic, not free-form). We keep expressivity by fixing the qualitative components (selecting the article, isolating the discriminating value, choosing a tolerance) at annotation time, verified against the primary source (the versioned article text); at scoring time the check is a regex or a numeric equality. Disagreement is thus resolvable by re-reading the corpus rather than re-prompting a judge.

\subsection{Parametric Knowledge Filter}

Following \citet{opsahlong2026officeqa}, we validate that the benchmark requires grounding (not parametric knowledge) via a zero-knowledge baseline: each candidate question is put to every evaluated LLM \emph{without} corpus or web access, and any question whose gold value is stated in at least one of four sampling draws by any model is dropped. We tighten \citeauthor{opsahlong2026officeqa}'s ``union over frontier models'' criterion to a union over \emph{all eleven evaluated models}, i.e., the five frontier systems, the substitute Gemini entry, and the five open-weight systems used in $C_{prod}$ (\S\ref{sec:experiments}), so the retained set is all-model-hard rather than only all-frontier-hard.\footnote{On the Qwen substitution between selection and evaluation, see note~$\ddagger$ of Table~\ref{tab:killer}.}

\paragraph{Current-version divergence filter.} A second, LLM-free filter enforces the premise of Condition~B (\S\ref{sec:experiments}): a regex check tests whether the gold date-anchored value still appears verbatim in the \emph{currently in-force} version of the article, and candidates whose value is unchanged are dropped. On the final scored set the current text lacks the gold value for \textbf{208 of the 209} questions (the single exception is kept as a within-set well-formedness control, \S\ref{ssec:killer-results}); where an article has no currently in-force version, divergence holds trivially and Condition~B retrieves nothing. We state the consequence explicitly: \textbf{Condition~B fails partly by construction}: the retained questions are screened so that the current-version text does not contain the gold value. The falsifiable content of the corresponding hypothesis is therefore its provenance and ceiling components, not the low strict score itself (see the reformulated $H_2$, \S\ref{sec:experiments}).

\subsection{Question Curation}

We are explicit about the manual and automatic components of the benchmark, a common concern for benchmark papers.

\paragraph{Questions: corpus-surfaced candidates, author-written questions.} The questions are produced by a two-stage pipeline, both stages released in the repository. First, candidate drift points are surfaced \emph{automatically} from the versioned corpus: \texttt{r3\_factory.py} scans version histories for value transitions (a threshold or rate that changes between consecutive versions) and \texttt{r3\_worksheet.py} emits them as a working sheet (article, transition date, before/after values). Second, every question, date anchor, canonical answer, and nugget is \emph{written by the authors}, who draft the question text, verify the value against the corpus version, and set the tolerance; the full set was then reviewed for correctness by a qualified French tax professional. A subset originates from an internal pool (Fiscal-FR-Bench v1) previously assembled by the authors.\footnote{Fiscal-FR-Bench v1 is a proprietary benchmark held by the authors. We release the R3 temporal track (221 questions, 209 in scoring scope), together with the R2 (43) and R4 (209) tracks, as Fiscal-FR-Bench v0 alongside this paper.}

\paragraph{Selection criteria.} The 209 scored questions (after the all-model-hard parametric filter and the answerable-scope audit) cover 33 CGI articles across seven fiscal sub-domains (Table~\ref{tab:distribution}): personal income tax (IR), local business and housing taxes (IFER, CFE, TH, TSE), betting and gaming, indirect taxes (VAT, tobacco, TV, mining), BIC/agricultural and cross-border, wage tax, and wealth tax (ISF). Each references an article whose content changed measurably over 1989--2025 (statutory drift). Consistent with the filter, we target \emph{obscure, non-rounded} parameters (indexed allowances, thresholds, per-installation IFER tariffs) rather than well-known headline rates (e.g., the corporate rate of art.~219 or the income-tax scale of art.~197) that frontier LLMs reliably memorize.

\paragraph{Frozen evaluation set.} The 209-question scored set is frozen prior to any retriever or reranker development (\texttt{killer\_qids\_v2.txt}); the answerable-scope audit flags twelve released questions as out of scope by a model-independent criterion --- four (art.~199~undecies~A, whose per-square-metre ceiling is revalued annually by statutory INSEE indexation and published via BOFiP; conservatively excluded at curation) plus eight caught in full-set review (curation errors or items ill-posed under their date anchor; itemized in the repository's \texttt{excluded\_qids\_scope.txt}) --- leaving $k{=}209$ scored out of 221 released. The fiscal reranker we additionally evaluated is fit only on articles \emph{disjoint} from those of this set; it yields no top-1 gain and is not part of the reported $C_{prod}$ pipeline.

\paragraph{Nuggets and answers: fully manual.} Each question is paired with two atomic required nuggets, manually annotated by the authors: a regex on the applicable article number and a numeric-with-tolerance match on the exact date-applicable value. Because the value nugget is the \emph{exact date-applicable} figure (numeric-with-tolerance), strict accuracy already requires the correct-date value, not merely a plausible article; version-selection quality is tracked \emph{separately} by the retrieval-side provenance score (\S\ref{ssec:nuggets}), so the two metrics are decoupled and neither inflates the other. The canonical answer is cross-checked verbatim against the version the nuggets point to (all 221 curated questions verified against the corpus) and reviewed for correctness by a qualified French tax professional. The jurisprudence links (\S\ref{ssec:linking}) serve only as weak supervision for R1/R4, never as ground truth for R3.

\paragraph{Parametric filter, in practice.} At selection time, no evaluated model recovered its date-applicable answer in any of four sampling draws; this is the sense in which the set is all-model-hard. At evaluation time, mean Condition~A strict accuracy is 3.0\% across the eleven models (cluster-bootstrap 95\% CI [1.4, 4.7]; per-model values in Table~\ref{tab:killer}). Residual variance stems from the Anthropic frontier models' default sampling temperature (genuinely diversified draws), the GPT-5.x reasoning stack's non-determinism even at temperature~0, and the remaining entries' effectively single deterministic draws at temperature~0, which make the filter conservative rather than lenient (a memorized answer would appear identically in all four draws and be caught).

\paragraph{Residual leakage.} Selection-time hardness does not freeze evaluation-time behavior: across the eleven models, at least one model produced the gold \emph{value} in Condition~A for 37 of the 209 questions (17.7\%; per-model counts 0--25, highest on GPT-5.5 (25) and Opus~4.7 (16), reflected in their higher Cond~A strict in Table~\ref{tab:killer}): sampling drift and value reconstruction, not memorization (a value memorized identically across all four draws would have been caught at selection). The set is thus all-model-hard under the models' evaluation configurations, and the B$\rightarrow$C comparison is unaffected, as all conditions share the configuration. Per-model counts: \texttt{stats\_clustered\_report.md}~(\S4).

\paragraph{Why 209 questions?}
The original R3 track was deliberately small ($k{=}35$; cf.\ LegalBench's 162 tasks, \citealp{guha2024legalbench}), designed to isolate the temporal-misgrounding phenomenon at minimal annotation cost. To respond to reviewer feedback on scale and to make room for eleven models (frontier + open-weight), we extended the track to 221 curated questions over the same annotation protocol (209 in scoring scope after the answerable-scope audit), spanning 33 articles instead of 10. Each additional question follows the same curation protocol (selection, version identification, canonical answer, two nuggets, corpus cross-validation) and survives the tightened all-model-hard filter. We now read Table~\ref{tab:killer} at both the condition-gap level (A/B $\ll$ C, unchanged from the original submission) and the model-ranking level (frontier vs.\ open-weight), which the larger $k$ makes statistically well-powered (\S\ref{ssec:killer-results}).

The per-article and per-sub-domain distribution of the scored set is given in
Table~\ref{tab:distribution} (Appendix~\ref{app:corpus}).

\section{Experiments}
\label{sec:experiments}

We run a \textbf{controlled three-condition experiment} measuring the accuracy gap between standard LLM/RAG approaches and our temporally-versioned retrieval on R3 questions.

\subsection{Controlled Experiment: Temporal Drift}
\label{ssec:killer}

\paragraph{Design.}
We evaluate three retrieval conditions on the frozen all-model-hard R3 subset ($k{=}209$ scored; Table~\ref{tab:distribution}), with eleven answer models: five frontier closed-API systems (Claude Opus 4.7, Claude Opus 4.8, Claude Sonnet 4.6, GPT-5.4, GPT-5.5), plus Gemini 2.5 Pro as a substitute frontier entry (Gemini 3 Pro Preview being rate-limited during the evaluation window; note~$\dagger$ of Table~\ref{tab:killer}), and five open-weight systems: Mistral Large 2407, Llama 4 Maverick, Qwen 3 235B, Gemma 3 27B, and GLM 5.2. The frontier set includes two Opus and two GPT-5 generations to test whether within-family generation gains close the temporal-misgrounding gap parametrically (they do not; \S\ref{sec:results}). The five open-weight systems, run via public inference endpoints (OpenRouter and Together AI), let us test whether temporal grounding depends on model scale or provider (it does not).

\begin{enumerate}
  \item \textbf{Condition A: LLM-only.} The model answers using parametric knowledge only: no retrieval, no web. A single evaluation draw per question; the four-draw probing applies to the selection-time parametric filter (\S\ref{sec:benchmark}; sampling configurations in the caption of Table~\ref{tab:killer}).
  \item \textbf{Condition B: RAG over current corpus.} The model retrieves over the static, current-version-only CGI corpus (representative of the majority of deployed legal RAG systems today). Article retrieval uses the gold \texttt{cid}; only the \emph{current} in-force version is exposed. Because B is handed the gold \texttt{cid}, it is charitable to the static baseline, so the reported A/B\,$\ll$\,C gap is a lower bound.
  \item \textbf{Condition C: RAG over our versioned corpus.} The model retrieves over our temporally-versioned corpus, where the version layer returns the article version applicable at the date in the question. We evaluate two retrievers that share this version layer:
  \begin{itemize}
    \item \textbf{$C_{or}$ (oracle):} article retrieval uses the gold \texttt{cid}, isolating the version-selection contribution (the ceiling once the right article is found).
    \item \textbf{$C_{prod}$ (end-to-end):} an end-to-end retriever finds \emph{both} the article and the version, with no oracle: the realistic deployment setting. The dense channel is a domain-adapted dense encoder over a chunked, \emph{multi-version} article index (three versions per \texttt{cid}: first, median, last by \texttt{date\_debut}, excluding stillborn \texttt{MODIFIE\_MORT\_NE} artefacts); the sparse channel is BM25 over the same chunks; the two are fused by reciprocal-rank fusion (no cross-encoder rerank in $C_{prod}$). The top-5 unique articles are handed to the version layer, which resolves the date-applicable version per \texttt{cid} before prompting the LLM.
  \end{itemize}
\end{enumerate}

\paragraph{Metric.}
Per-question metrics (\emph{coverage}, \emph{strict}, \emph{provenance}) are as defined in \S\ref{ssec:nuggets}; per-condition scores are means over the $k{=}209$ scored questions, either per model or pooled over the eleven models.

\paragraph{Hypotheses.} We test four falsifiable predictions, each tied to a condition:
($H_1$)~A is low on strict accuracy (near-absent parametric knowledge of date-specific values), uniformly across scales and providers, though coverage stays non-trivial since article-id nuggets are easy;
($H_2$)~B retrieves the date-applicable version $0\%$ of the time and stays below $10\%$ strict. The near-zero \emph{strict} score is largely a \emph{construction check}, since the current-version divergence filter (\S\ref{sec:benchmark}) screens retained questions so the current text does not contain the gold value; the \emph{falsifiable} content of $H_2$ is the $0\%$ provenance (structural: a single-version index cannot hold the version) and the $<10\%$ ceiling, which a model could in principle beat by re-deriving the historical value from the current text (e.g., un-indexing a revalorized threshold);
($H_3$)~$C_{or}$ closes most of the gap ($>80\%$ strict, $100\%$ provenance), showing version selection (not corpus completeness or model size) is the bottleneck;
($H_4$)~a realistic retriever feeding the top-5 date-applicable versions recovers essentially the $C_{or}$ ceiling, the residual gap being a recall@5 ceiling on a few niche queries, so the lever is recall, not top-1 reranking.

\section{Results}
\label{sec:results}

\subsection{Corpus Quality}

\paragraph{Versioning depth.} Table~\ref{tab:corpus-stats} reports the distribution of article-versions per code: a mean of 4.02 versions per \texttt{cid}, with the CGI principal article 81 (tax-exempt income), repeatedly amended by finance laws, holding 94 distinct historical versions. This depth, the very structure a static index collapses (\S\ref{ssec:structural}), is what the controlled experiment exploits. Quality of the auxiliary jurisprudence-linking dataset (precision, recall, gold-standard audit) is reported in Appendix~\ref{app:linking}.

\subsection{Controlled Experiment Results}
\label{ssec:killer-results}

\begin{table*}[t]
\centering
\small
\setlength{\tabcolsep}{4pt}
\begin{tabular}{lccccccccccc}
\toprule
 & \multicolumn{4}{c}{Coverage} & \multicolumn{4}{c}{Strict} & \multicolumn{3}{c}{Provenance}\\
\cmidrule(lr){2-5}\cmidrule(lr){6-9}\cmidrule(lr){10-12}
Model & A & B & $C_{or}$ & $C_{prod}$ & A & B & $C_{or}$ & $C_{prod}$ & B & $C_{or}$ & $C_{prod}$\\
\midrule
\multicolumn{12}{l}{\emph{Frontier (closed API)}}\\
Opus 4.7                 & 53.8 & 55.3 & 100.0 & 99.8 & 7.7 & 10.5 & \textbf{100.0} & \textbf{99.5} & 0 & 100 & 99\\
Opus 4.8                 & 53.3 & 51.4 & 100.0 & 99.5 & 6.7 & 3.3 & \textbf{100.0} & \textbf{99.0} & 0 & 100 & 99\\
Sonnet 4.6               & 49.5 & 50.0 & 100.0 & 99.5 & 1.0 & 1.4 & \textbf{100.0} & \textbf{99.0} & 0 & 100 & 99\\
GPT-5.4                  & 50.7 & 51.2 & 100.0 & 99.5 & 1.4 & 2.4 & \textbf{100.0} & \textbf{99.0} & 0 & 100 & 99\\
GPT-5.5                  & 56.0 & 54.1 & 100.0 & 99.8 & 12.0 & 8.1 & \textbf{100.0} & \textbf{99.5} & 0 & 100 & 99\\
Gemini 2.5 Pro$^{\dagger}$ & 46.2 & 49.5 & 99.0 & 99.0 & 1.4 & 0.5 & \textbf{99.0} & \textbf{98.6} & 0 & 100 & 99\\
\midrule
\multicolumn{12}{l}{\emph{Open-weight}}\\
Mistral Large 2407       & 50.2 & 50.7 & 99.8 & 99.0 & 0.5 & 1.4 & \textbf{99.5} & \textbf{98.1} & 0 & 100 & 99\\
Llama 4 Maverick         & 48.1 & 50.2 & 99.0 & 98.6 & 0.0 & 0.5 & \textbf{98.1} & \textbf{97.1} & 0 & 100 & 99\\
Qwen 3 235B$^{\ddagger}$ & 46.7 & 49.3 & 97.4 & 98.6 & 1.0 & 0.5 & \textbf{94.7} & \textbf{97.1} & 0 & 100 & 99\\
Gemma 3 27B              & 42.8 & 48.6 & 99.8 & 97.4 & 0.0 & 0.5 & \textbf{99.5} & \textbf{95.2} & 0 & 100 & 99\\
GLM 5.2                  & 49.0 & 50.2 & 99.8 & 99.5 & 1.0 & 1.0 & \textbf{99.5} & \textbf{99.0} & 0 & 100 & 99\\
\midrule
Mean (11 models)         & 49.7 & 51.0 & 99.5 & 99.1 & 3.0 & 2.7 & \textbf{99.1} & \textbf{98.3} & 0 & 100 & 99\\
\bottomrule
\end{tabular}
\caption{Results on the $k{=}209$ scored all-model-hard R3 subset (questions no
evaluated model, frontier or open, answered from parametric knowledge across
four sampling draws at selection time; 221 curated questions are released, twelve
flagged out of the answerable scope). \textbf{A}: LLM-only, a single
evaluation draw per question (the three Anthropic frontier models use their
default sampling temperature; GPT-5.x's reasoning stack is non-deterministic
even at temperature~0; the remaining six entries (five open-weight and
Gemini 2.5 Pro) run at temperature~0, effectively deterministic). \textbf{B}: RAG over the static current-version corpus
(gold article, current version). \textbf{$C_{or}$}: oracle version selection
(gold article, date-applicable version): the single-article ceiling.
\textbf{$C_{prod}$}: end-to-end retriever feeding the \emph{top-5}
date-applicable versions, no oracle (retriever details in
\S\ref{sec:experiments}). Coverage = mean nugget fraction; Strict = both
required nuggets hit (correct article and exact value); Provenance = \% of
questions whose date-applicable gold version is among the retrieved top-5.
All conditions share an identical labeled-article prompt at an 8{,}000-char/article
query-windowed budget (value-preserving: the date-applicable gold value lies
within the served window for all 209 scored questions), so $C_{or}$ and
$C_{prod}$ differ only in the number of retrieved articles. All numbers are
percentages; $k{=}209$; cluster-aware 95\% bootstrap
CIs (articles resampled) and per-model exact McNemar/Wilcoxon tests are
reported in \S\ref{ssec:significance}.
$^{\dagger}$Gemini 3 Pro Preview was rate-limited during the extended evaluation
window; we report Google's next-tier available model (Gemini 2.5 Pro) as a
substitute frontier entry. Conditions A/B/$C_{or}$ ran on the native AI~Studio
endpoint; $C_{prod}$ ran via OpenRouter after the native per-day quota was
exhausted. Its mandatory reasoning stack is non-configurable, so Cond~A is a
single reasoning-conditioned draw. The Gemini 2.5 Pro and GLM 5.2 entries
include a small number of provider-side empty responses scored as failures
(22 in total across conditions, a conservative direction; audit in the
repository, \texttt{parametric\_filter\_empty\_audit.md}).
$^{\ddagger}$Qwen 2.5 72B (used at selection time for the parametric filter)
was retired from serverless inference by every accessible provider during the
evaluation window; we substitute Qwen 3 235B (Together AI). A larger, more
recent model could in principle recover \emph{more} questions parametrically;
empirically its Cond~A strict is 0.0\%, so the probe remains all-model-hard.}
\label{tab:killer}
\end{table*}

On the $k{=}209$ all-model-hard subset, parametric knowledge (A) is uniformly low across all eleven models (3.0\% mean strict, cluster-bootstrap 95\% CI [1.4, 4.7]; Table~\ref{tab:killer}), by construction of the filter, tightened from ``no frontier model recovers'' (\citealp{opsahlong2026officeqa}) to ``no evaluated model recovered across four sampling draws at selection time'', which excludes any question a model happened to memorize (residual evaluation-time drift is quantified in \S\ref{sec:benchmark}). Static-corpus RAG (B), the setting deployed by most legal-AI products today, does not improve on the LLM-only baseline (2.7\% mean strict, cluster-bootstrap 95\% CI [1.3, 4.8], statistically indistinguishable from A) and retrieves the date-applicable version in 0\% of cases: it fails not silently but \emph{confidently}, grounding on a real, well-formed, but inapplicable version (both are construction checks rather than surprising measurements; \S\ref{sec:benchmark} and $H_2$, \S\ref{sec:experiments}).

\paragraph{Control condition (well-formedness).} Two same-set controls rule out ill-posed questions or a broken pipeline as an explanation for B's failure. First, under the \emph{identical} labeled-article prompt and pipeline, oracle version selection ($C_{or}$, which changes only the served \emph{version} of the same article) reaches \textbf{99.1\%} mean strict on the same 209 questions (Table~\ref{tab:killer}): the questions are answerable and the model extracts the value correctly \emph{when handed the date-applicable version}, so B's near-zero score is not ill-posedness or a reading failure. Second, the current-version text lacks the gold date-anchored value for \textbf{208 of the 209} scored questions (the divergence premise of Condition~B, \S\ref{sec:benchmark}); on the single non-drifted exception, whose value \emph{is} present in the current text, static B retrieves it correctly. B's failure on the drifted set is therefore version drift, not a broken pipeline.

\paragraph{The operative result: end-to-end retrieval.}
$C_{prod}$, the end-to-end retriever, is given no oracle (it must locate both the article and its date-applicable version from the query alone) and reaches \textbf{98.3\% mean strict} (cluster-bootstrap 95\% CI [95.9, 99.6]); all eleven models cross 95\% in point estimates (95.2--99.5\%; every per-model clustered CI stays above 89\%, \S\ref{ssec:significance}), and the gold version is in the top-5 for 99\% of questions (vs.\ 0\% for B). Holding the versions is necessary but not sufficient: an oracle-article ablation ($C_{or}$, gold \texttt{cid}, leaving only version selection) reaches 99.1\% mean (CI [98.4, 99.9]) at 100\% provenance. The $C_{or}\rightarrow C_{prod}$ gap is only 0.8 points, concentrated on \emph{first-stage recall} rather than version selection: the sole provenance miss is art.~1417 (2 questions whose date-applicable version falls outside the retrieved top-5). Once the corpus is versioned the bottleneck is article recall, not date resolution; a cross-encoder reranker adds nothing on this niche set.

\paragraph{Significance (cluster-aware).}
\label{ssec:significance}
The protocol is paired, with two dependencies a pooled iid analysis would ignore: the eleven models answer the \emph{same} questions, and questions cluster by article (33 clusters; 34 on art.~1466~A alone). We therefore treat the \emph{model} as the unit of inference and bootstrap by resampling \emph{articles}, not questions (10{,}000 iterations, seed 42). The B$\rightarrow C_{or}$ and B$\rightarrow C_{prod}$ gains are significant \emph{for every model individually} (exact two-sided McNemar on per-question strict outcomes, 11 tests, all $p<10^{-55}$), so the conclusion does not depend on pooling; per-model Wilcoxon tests on coverage agree, while the $C_{or}\rightarrow C_{prod}$ coverage difference is small and not uniformly significant, consistent with the 1\% recall@5 ceiling. Full per-model tests are in the repository's \texttt{stats\_clustered\_report.md}. Cluster-bootstrap 95\% CIs for pooled strict: A 3.0~[1.4, 4.7], B 2.7~[1.3, 4.8], $C_{or}$ 99.1~[98.4, 99.9], $C_{prod}$ 98.3~[95.9, 99.6]; every per-model $C_{prod}$ lower bound stays above 89\% (the claim holds under clustering, not only as a point estimate). Leave-one-article-out on pooled $C_{prod}$ strict spans 98.1\% (dropping art.~1519~A) to 99.2\% (dropping art.~1417): no single article carries the result. A single-question walkthrough (art.~219 CGI) is in App.~\ref{app:case-study} (Fig.~\ref{fig:conditions}).

\paragraph{Windowing policy.}
All conditions serve a \emph{value-preserving, query-windowed} extract at an
8{,}000-char/article budget: for each retrieved article we keep the head plus
the passages most relevant to the query, capped at 8{,}000 characters. A
pre-scoring check confirms the date-applicable gold value lies within the served
window for all 209 scored questions, so no condition is handed an extract that
cannot contain the answer, and the budget affects B, $C_{or}$, and $C_{prod}$
symmetrically. This resolves the head-truncation artifact of the original
6{,}000-char cap, under which the gold value of a handful of long articles
(e.g., arts.~158, 1605~bis, 261, 83) fell beyond the window. Uniform full-text
prompting is not preferable: it degrades $C_{or}$ on long articles through
long-context distraction (ablation scripts in the repository).

\paragraph{Failure-mode distribution.}
\label{ssec:failure-distribution}
Condition B retrieves the current version in 100\% of cases
(Table~\ref{tab:killer}, provenance), so \textbf{100\% of B-condition errors are
current-law substitution}, the dominant mode by construction of a static index.

\section{Conclusion}
\label{sec:conclusion}

We identified \textbf{temporal misgrounding}, the systematic retrieval and citation of the currently-in-force version of a legal article when the question requires an earlier or future version, as a structural failure mode of legal RAG: stable identifiers, date-dependent correctness, and cross-version semantic similarity meeting single-version indexing, no date-conditioning, and parametric recency bias. To quantify it we built \textbf{FiscalQA Pro}: a temporally-versioned corpus of 32{,}436 CGI/LPF article-versions over 93 years (plus 69{,}208 auxiliary version-aware jurisprudence links) and an R3 temporal-reasoning track of 209 scored, expert-reviewed, all-model-hard questions across 33 CGI articles (221 released), scored deterministically via nuggets.

Across eleven models, our controlled experiment shows LLMs fail on temporally-grounded questions with or without static-corpus RAG (3.0\% and 2.7\% pooled mean strict; static RAG retrieves the date-applicable version 0\% of the time), while our end-to-end retriever, conditioning retrieval on the query date with no oracle, reaches 98.3\% mean strict (all eleven models cross 95\% in point estimates), an oracle-article ablation placing the ceiling at 99.1\% and the residual 0.8-point gap in first-stage recall rather than version selection. We argue legal QA should be reframed as a \emph{temporally-indexed retrieval problem}, and release the corpus, benchmark, model responses, and pipeline code (the fine-tuned encoder weights excepted; Appendix~\ref{app:repro}).

\section{Limitations and Future Work}
\label{sec:future}

\paragraph{Scope and civil-law generalization.}
FiscalQA Pro covers French tax law only (CGI, its four annexes, and the LPF).
We argue, however, that temporal misgrounding is \emph{structural to civil-law
statutory retrieval} rather than a French artifact: any corpus amended in place
and versioned over time exhibits the same conjunction of properties
(\S\ref{sec:problem}), and several jurisdictions expose versioned statutory APIs
with explicit validity dates analogous to Légifrance (Fedlex, Gesetze-im-Internet,
Justel, Légilux; \citealp{fedlex,gesetzeiminternet,justel,legilux}). Our method
layer is jurisdiction-agnostic; only the data layer is. The concurrent German
study of \citet{prior2026oldfriend} finds the same phenomenon---independent
evidence of generalization; a Swiss Fedlex replication is the next step.

\paragraph{Question scale.}
The R3 track contains 221 curated questions
(209 in scoring scope), up from 35 in the original submission
(\S\ref{sec:benchmark}). The extended set is
still smaller than LegalBench (162 tasks; \citealp{guha2024legalbench}) or
KARLBench (2{,}000+ questions): a deliberate trade against synthetic or
LLM-authored scaling, since concentrated, expert-reviewed, all-model-hard
difficulty is the right regime for measuring temporal misgrounding
(cf.\ OfficeQA Pro's 133 questions).

\paragraph{Failure modes evaluated.}
Our single-anchor R3 set exercises only \emph{current-law substitution}; the other
three taxonomy modes (\S\ref{ssec:taxonomy}) require future-effective (\texttt{VIGUEUR\_DIFF}),
``in its version prior to law X,'' and two-version-comparison questions, which we
leave to v1; the corpus's \texttt{VIGUEUR\_DIFF} versions, extending to 2031,
already enable the future-law-leakage mode via questions about \emph{future}
legal states fixed by enacted-but-deferred legislation. Finally, beyond
``apply the law'': R1--R3 target rule application, not \emph{legal
interpretation}; R4 is a first step toward interpretive benchmarks.


\section*{Impact Statement}
This work studies a reliability failure of legal AI, temporal misgrounding, and
provides a benchmark and methodology to measure and mitigate it. Improving the
temporal correctness of legal question answering can reduce confidently-wrong
outputs in high-stakes settings (tax compliance, legal research). The corpus and
benchmark are built entirely from public legislation and case law; no personal
data is introduced. As with any legal-AI tool, outputs should be verified by a
qualified professional and not treated as legal advice.

\bibliography{references}
\bibliographystyle{icml2026}

\appendix
\section{Corpus and Benchmark Statistics}
\label{app:corpus}

Table~\ref{tab:corpus-stats} breaks the corpus down by code: versions,
distinct articles (\texttt{cid}), and average versioning depth.
Table~\ref{tab:distribution} gives the distribution of the $k{=}209$ scored R3
questions across CGI articles and fiscal sub-domains.

Two endpoint caveats on the 93-year span. The 1938 origin is carried by a
single LPF article-version (art.~L28) whose \texttt{date\_debut} records the
entry into force of the predecessor provision it codifies (d\'ecret-loi of
1~June~1938); the LPF itself dates from 1981--82. The 2031 endpoint excludes
20 rows carrying L\'egifrance placeholder dates (2222-02-22, 2999-01-01),
which the version-selection layer never serves.

\begin{table}[h!]
\centering
\small
\begin{tabular}{lrrr}
\toprule
\textbf{Code} & \textbf{Versions} & \textbf{CIDs} & \textbf{V/CID} \\
\midrule
CGI principal             & 21{,}040 & 3{,}696 & 5.69 \\
CGI annexe III            & 3{,}801  & 1{,}467 & 2.59 \\
LPF                       & 3{,}031  & 1{,}095 & 2.77 \\
CGI annexe II             & 2{,}390  & 1{,}005 & 2.38 \\
CGI annexe IV             & 1{,}879  & 705     & 2.67 \\
CGI annexe I              & 295      & 108     & 2.73 \\
\midrule
\textbf{Total}            & \textbf{32{,}436} & \textbf{8{,}076} & \textbf{4.02} \\
\bottomrule
\end{tabular}
\caption{Tax legislation corpus statistics. V/CID = average number of versions per article (constant identifier).}
\label{tab:corpus-stats}
\end{table}

\begin{table}[h!]
\centering
\footnotesize
\setlength{\tabcolsep}{3pt}
\begin{tabular}{p{3.2cm}p{2.6cm}rc}
\toprule
Sub-domain & Principal arts.\ (\# Q) & \# & Span \\
\midrule
Local business \& housing taxes (IFER, CFE, TH, TSE)
  & 1466\,A~(34), 1519\,A~(20), 1586\,nonies~(17), 1519\,HA~(16), 1647\,D~(4), 1414\,A~(4), 1609\,C~(4), 1417~(2), 1414\,B~(1)
                                                                                              & 102 & 1990--2024 \\
Personal income tax (IR)
  & 156~(16), 196\,B~(5), 168~(5), 158~(4), 157\,bis~(4), 204\,H~(4), 200~(2), 199\,decies\,H~(2), 83~(2), 199\,sexies~(1), 5~(1)
                                                                                              & 46 & 1989--2024 \\
Betting \& gaming
  & 302\,bis\,ZI~(16), 302\,bis\,ZG~(9)                                        & 25 & 2012--2025 \\
Indirect taxes (VAT, tobacco, TV, mining)
  & 568~(7), 1587~(7), 261~(3), 302\,bis\,ZC~(3), 1605\,bis~(3)      & 23 & 1996--2025 \\
BIC, agricultural \& cross-border
  & 50-0~(3), 182\,A~(1), 1679\,A~(1), 73~(1)              & 6  & 1991--2025 \\
Wage tax
  & 231~(5)                                                             & 5  & 2002--2019 \\
Wealth tax (ISF)
  & 885\,H~(2)                                                          & 2  & 2010--2014 \\
\midrule
\textbf{Total} & \textbf{33 articles, 7 sub-domains} & \textbf{209} & \textbf{1989--2025} \\
\bottomrule
\end{tabular}
\caption{Distribution of the $k{=}209$ scored all-model-hard temporal-reasoning
(R3) questions across CGI articles and fiscal sub-domains (the twelve released
questions flagged out of the answerable scope---art.~199~undecies~A and eight
review exclusions---are not shown). The set spans seven tax sub-domains and 36 years of statutory drift
(1989--2025), including the franc-to-euro transition. Articles are listed in
decreasing order of question-count within each sub-domain (parenthetical
count).}
\label{tab:distribution}
\end{table}

\section{Version-Aware Jurisprudence Linking}
\label{app:linking}

Alongside the versioned legislation, we release a dataset linking tax-related
court decisions to the \emph{specific article version} applicable at the date of
decision. This resource is auxiliary to the temporal-misgrounding study (it is
not used by the controlled experiment of \S\ref{sec:experiments}); we provide its
full construction and quality analysis here.

\subsection{Construction}

\paragraph{Source decisions.} We work with four sources of French tax-related
case law: \texttt{arianeweb\_decisions} (Conseil d'État, 51{,}564 decisions),
\texttt{inca} (Cour de cassation, 300{,}383), \texttt{judilibre\_decisions}
(Cour de cassation with structured visa, 265{,}891), and
\texttt{decisions\_unified} (an aggregation of the above plus EU and sectoral
authorities, 663{,}748). To avoid double-counting between the aggregation and
its underlying sources, we report metrics by individual source. We restrict each
source to a fiscal subset using either a PostgreSQL \texttt{tsvector} query (for
\texttt{decisions\_unified}) or pattern matches on tax-relevant terms (for the
others); the resulting subsets total 61{,}835 decisions.

\paragraph{Citation extraction and version selection.} For each fiscal decision,
we apply a regular-expression extractor sensitive to French tax-citation
conventions: numbered articles (e.g., \texttt{article 209 B}), LPF prefixes
(\texttt{L.\ 16 B}, \texttt{R.\ 281-1}), and Latin ordinals (\texttt{bis},
\texttt{ter}, \texttt{quater}, \dots). Candidates are filtered by a two-tier
disambiguation: (i) a proximity veto rejecting matches whose immediate
($\pm 70$ chars) right-context attaches them to another source (e.g.,
\texttt{du Code civil}, \texttt{de la loi n°\ X}), and (ii) a fiscal-context
requirement in a $\pm 100$-char window.\footnote{For ambiguous one- to
three-digit numbers, we require a \emph{strong} fiscal context (explicit mention
of CGI/LPF). For numbers with suffix or LPF prefix, a weaker fiscal context
suffices.} Each surviving candidate is resolved against an in-memory article
index; we then select the article \emph{version} valid at the decision's date by
intersecting against \texttt{date\_debut} and \texttt{date\_fin} ranges. Where
structured \texttt{visa} fields are available (Judilibre), we tag the link as
\texttt{VISE} (formally cited as legal basis); all other links are tagged
\texttt{CITE}.

\paragraph{Output.} Linking produces 69{,}208 version-aware links across 32{,}034
distinct decisions. Of these, 97\% point to a version whose validity period
strictly includes the date of decision; the rest fall back to the most recently
active version at decision time. The linked decisions span five decades
(the 2000s most represented; Table~\ref{tab:decade-distribution}), enabling
longitudinal evaluation. The decade distribution and the validity-inclusion
rate are computed on the underlying decision tables, reconstructible from the
public sources via the released extraction scripts; the released linking
dataset ships the resolved links themselves.

\begin{table}[t]
\centering
\small
\begin{tabular}{lrr}
\toprule
\textbf{Decade} & \textbf{Decisions} & \textbf{Links} \\
\midrule
Before 1990  & 7{,}629  & 15{,}953 \\
1990--1999   & 6{,}995  & 13{,}774 \\
2000--2009   & 7{,}679  & 17{,}182 \\
2010--2019   & 4{,}991  & 10{,}588 \\
2020--2026   & 4{,}740  & 11{,}711 \\
\midrule
\textbf{Total} & \textbf{32{,}034} & \textbf{69{,}208} \\
\bottomrule
\end{tabular}
\caption{Distribution of version-aware jurisprudence links and linked decisions
by decade. Each decision is bucketed by its decision date where available, else
by the earliest applicable-version start date (for the 25\% of links from
aggregated \texttt{decisions\_unified} records whose decision-date field is
unpopulated in this snapshot); totals match the 69{,}208 links across 32{,}034
decisions.}
\label{tab:decade-distribution}
\end{table}

\subsection{Quality Audit}
\label{app:audit}

\paragraph{Precision.} We sample 100 links proportionally stratified by source
(seed=42, reproducible) and audit them with a two-stage auto-annotator followed
by manual review of low-confidence cases. The auto-annotator first checks the
immediate right-context ($\pm 60$ chars after the article number) for explicit
CGI/LPF mention (high-confidence true positive) or competing source attachment
(high-confidence false positive); cases with neither signal are checked in a
wider $\pm 300$-char window. We measure \textbf{98--99\% precision} on this
stratified sample. The single residual false positive is a header-extraction
artifact (an article number in a court-arrêt metadata header without fiscal
context). Precision is uniform across sources (Conseil d'État: 100\%, Cour de
cassation: 92\%, Judilibre: 100\%, \texttt{decisions\_unified}: 99\%) and across
article-number types (alphabetic-suffix, Latin-ordinal, short-numeric,
LPF-prefixed: all $\geq 97\%$).

\paragraph{Failure modes.} The remaining 1--2\% of false positives fall into four
patterns: (a) cross-code number collisions (e.g., \texttt{article 1382} of the
Civil Code vs.\ the CGI), (b) attachment to a numbered law or convention rather
than a code, (c) extraction from non-substantive metadata (decision headers),
and (d) ambiguous coreferences such as ``le même code.'' Patterns (a) and (b)
are largely addressed by the proximity veto; (c) and (d) motivate the NER-based
fine-tuning discussed below (\emph{Limitations and next step}).

\paragraph{Cross-validation against Judilibre visa.} Of the 907 fiscal
Judilibre decisions, 843 (93\%) receive at least one link (decision-level
recall). On the 485 decisions whose visa is parseable as referring to CGI/LPF,
article-level recall on visa-matched articles is 37.6\% after fuzzy stem
matching. The remaining gap is partially explained by (i) coarser visa labels
(e.g., \texttt{L16}) vs.\ finer textual citations (\texttt{L.\ 16 B}), and
(ii) articles cited in the decision's reasoning that are not formally part of the
visa.

\paragraph{Manual gold-standard recall.} All CGI/LPF articles cited in a
stratified random sample of 50 fiscal decisions were exhaustively annotated
(read on full text, not the truncated extract), yielding 116 article
references across 34 decisions (16 decisions cite no nominative CGI/LPF article).
Comparing the pipeline's links against this ground truth, with article-identity
normalization applied symmetrically to both sides (alphabetic suffixes and Latin
ordinals kept; paragraph/alinea markers stripped), we measure \textbf{88.6\%
article-level precision}, \textbf{53.4\% article-level recall}, and \textbf{97\%
decision-level recall} (33/34 decisions receive at least one correct link).
Recall is highest on structured jurisdictional sources (Judilibre 71\%,
\texttt{decisions\_unified} 55\%, Arianeweb 53\%) and lowest on \texttt{inca}
(48\%). The residual false-negatives concentrate in (i) citations appearing deep
in long decisions (20--45k characters, beyond the span processed by the
extractor) and (ii) LPF procedural articles (\texttt{L.}/\texttt{R.} series);
both motivate the NER-based extractor below. This article-level
measurement complements the link-level precision audit above: the pipeline finds
\emph{at least one} grounding article in nearly every decision, and about half of
\emph{all} cited articles.

\paragraph{Limitations and next step.}
Recall is uneven: 83--93\% on the three jurisdictional sources but only 39\% on
the noisier aggregated \texttt{decisions\_unified} table, with 1--2\% residual
false positives (cross-code collisions, header artifacts, ambiguous
coreferences). A fine-tuned legal NER extractor, weakly supervised by the
69{,}208 regex links plus the 50-decision gold standard, is the natural next
step to close most of this gap.

\section{Case Study: Temporal Misgrounding on Article 219 CGI}
\label{app:case-study}

\begin{figure}[t]
\centering
\newcommand{\cmark}{\textcolor{green!45!black}{\checkmark}}
\newcommand{\xmark}{\textcolor{red!70!black}{\boldmath$\times$}}
\begin{tikzpicture}[
  >=Latex, node distance=2.6mm,
  row/.style={draw, rounded corners=2pt, align=left, text width=64mm, inner sep=4pt, font=\scriptsize},
  none/.style={row, dashed, fill=gray!8},
  wrong/.style={row, fill=red!8, draw=red!55},
  good/.style={row, fill=green!10, draw=green!50!black},
]
\node[draw, rounded corners=2pt, fill=yellow!15, text width=64mm, align=center, inner sep=4pt, font=\scriptsize] (q)
  {\textbf{Q:} standard corporate income tax rate for fiscal years opening on/after 1~Jan 2018?\\(gold: $33\tfrac{1}{3}\%$, art.~219 CGI)};
\node[none, below=of q] (a)
  {\textbf{A}~LLM-only \,(no retrieval)\hfill $25\%$~\xmark};
\node[wrong, below=of a] (b)
  {\textbf{B}~static RAG: art.~219 \textbf{v.\,2026} ``25\,\%''\hfill $25\%$~\xmark\\\mbox{}\hfill{\itshape confidently wrong}};
\node[good, below=of b] (c)
  {\textbf{C}~versioned (ours): art.~219 \textbf{v.\,2018} ``33 1/3\,\%''\hfill $33\tfrac{1}{3}\%$~\cmark};
\draw[->] (q) -- (a);
\draw[->] (a) -- (b);
\draw[->] (b) -- (c);
\end{tikzpicture}
\caption{The three retrieval conditions on a temporal-drift question
(art.~219 CGI, fiscal year 2018). \textbf{B} retrieves a real but inapplicable
version, the current consolidated version (2026 consolidation; its 25\% rate
has been unchanged since 2022), and is \emph{confidently wrong}; \textbf{C}
conditions retrieval on the question's date, returns the 2018 version, and
grounds the correct answer. Same LLM, same corpus; only version selection
differs.}
\label{fig:conditions}
\end{figure}
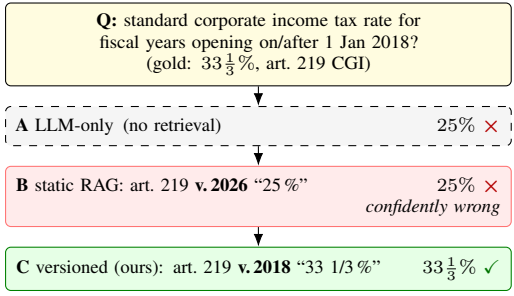

Figure~\ref{fig:conditions} walks a single R3 question end-to-end through the
three conditions. The question asks for the standard corporate income tax rate
for fiscal years opening on or after 1 January 2018 (gold: art.~219 CGI, version
\texttt{LEGIARTI000036431672} valid 2018-01-01, ``\dots fix\'e \`a 33 1/3\,\%'';
nuggets \texttt{\{art\_219\_CGI, 33\_one\_third\_pct, version\_2018,
rate\_general\_regime\}}). LLM-only (A) and static-corpus RAG (B) both answer
``25\%'' (2/4 nuggets): A from parametric recency bias, B by grounding on a
\emph{real but inapplicable} current consolidated version (2026 consolidation;
the 25\% rate itself has been unchanged since 2022): the most
insidious mode, since provenance is preserved but correctness is not.
Conditioning retrieval on the question's date (C) returns the 2018 version and
yields the correct ``$33\tfrac{1}{3}\%$'' (4/4). Same LLM, same corpus, only
version selection differs, and the gap moves from 50\% to 100\% on this
question. (We use art.~219 here only for clarity; the \emph{scored} set targets
less-memorized articles, Table~\ref{tab:distribution}.)

\section{Reproducibility}
\label{app:repro}

We list explicitly what is released.\footnote{\url{https://github.com/rosecymbler/fiscal-fr-bench}}
\textbf{Data (CC-BY-4.0, with Légifrance/Judilibre--Etalab~2.0 attribution;
SHA256 checksums included):} the versioned corpus (32{,}436 article-versions,
Parquet); the linking dataset (69{,}208 version-aware citations); the benchmark
(R3, 221 questions with 209 in scoring scope, plus the R2/R4 tracks, with
nuggets and ground-truth answers); the 100-link stratified audit sample and the
50-decision gold standard; all model responses behind Table~\ref{tab:killer}
(the query-windowing ablation of \S\ref{ssec:killer-results} is re-runnable
via the released scripts). \textbf{Code (MIT):} chrono extraction, linking,
audit and gold-standard sampling, the experiment harness for the reproducible
conditions (all condition prompts and the 8{,}000-char/article query-windowed
budget are verifiable in \texttt{run\_conditions.py}; conditions A, B and
$C_{or}$ run with \texttt{-{}-retriever oracle}), the deterministic scorer, and
the cluster-aware statistics (\texttt{stats\_clustered.py}). \textbf{Not
released:} the end-to-end $C_{prod}$ retriever---its domain-adapted encoder
\emph{weights}, multi-version index, and inference code, all proprietary.
$C_{prod}$ is therefore not reproducible from the released artifacts; we release
its per-question model responses so that the reported $C_{prod}$ scores remain
independently re-verifiable with the deterministic scorer. Between the original submission and this camera-ready
pass, the article index was revised from a single ``longest'' version per
\texttt{cid} to three representative versions per \texttt{cid} (first, median,
last by \texttt{date\_debut}, excluding stillborn \texttt{MODIFIE\_MORT\_NE}
artefacts), holding the embedding model, BM25 hyper-parameters, chunker, and
RRF fusion constant. On the associated contamination question: the encoder's
fine-tuning pairs overlap the benchmark on five released articles at the
article level---art.~150~U, one of the five, is flagged out of the answerable
scope, leaving 28 affected \emph{scored} questions on four of the 33 scored
articles---but
\emph{zero} of those questions' gold values appear
in any training passage (the pairs hold current-version texts; the benchmark targets historical values); the check, with its re-runnable script, is available from the authors on request (the
reranker's training articles are disjoint from the benchmark's, as stated in
\S\ref{sec:benchmark}; the encoder needed this finer value-level check). The
benchmark files embed a BIG-bench-style \emph{canary GUID}
so that future training-set contamination is detectable. All extraction is
reproducible end-to-end from public Légifrance and Judilibre data (free API
access via PISTE and the Cour de cassation portal). Random seeds are fixed (42
for the audit sample, 2026 for the gold standard, 42 for the scorer and
clustered bootstraps). Provider-specific evaluation settings are documented in
code and in the repository README (SDK version pins; GLM 5.2's reasoning
disabled via OpenRouter's \texttt{extra\_body} flag; GPT-5.5 called with
\texttt{reasoning\_effort="none"} for parity with GPT-5.4 on closed-book
probing). The repository also contains README, METHODOLOGY, DATA\_SCHEMA, and a
SPEC\_GOLD\_STANDARD document specifying the annotation protocol followed by
the authors.

\section{Control Condition (Well-Formedness)}
\label{app:control}

The well-formedness of Condition~B's failure is established \emph{within} the
scored set (\S\ref{ssec:killer-results}), without a separate control split. Two
facts jointly rule out ill-posed questions or a broken pipeline. (i)~Under the
identical labeled-article prompt and retrieval pipeline, changing only the
served \emph{version} of the same article (oracle version selection, $C_{or}$)
lifts pooled strict accuracy from 2.7\% (Condition~B) to \textbf{99.1\%} on the
same 209 questions (Table~\ref{tab:killer}): the questions are answerable and the
model extracts the value correctly \emph{when handed the date-applicable
version}. (ii)~The current-version text lacks the gold date-anchored value for
\textbf{208 of the 209} scored questions (\S\ref{sec:benchmark}); on the single
non-drifted exception the value \emph{is} present and static Condition~B
retrieves it correctly. Condition~B's near-zero strict score is therefore
version drift, not question difficulty or a pipeline defect.

\end{document}